\documentclass{tlp}
\usepackage{graphicx}

\newcommand{\clpr}{CLP({\ensuremath{\cal R}})}
\newcommand{\eclipse}{ECL$^i$PS\ensuremath{^e}}
\newcommand{\Mop}{{\ensuremath{\cal M}}}    
\newcommand{\mop}{{\ensuremath{m}}}         
\newcommand{\Mpr}{{\ensuremath{\cal N}}}    
\newcommand{\mpr}{{\ensuremath{n}}}         
\newcommand{\Ope}{{\ensuremath{\bf A}}}     
\newcommand{\ope}{{\ensuremath{a}}}         
\newcommand{\Nope}{{\ensuremath{N_a}}}   
\newcommand{\Pre}{{\ensuremath{\bf P}}}     
\newcommand{\pre}{{\ensuremath{p}}}         
\newcommand{\Npre}{{\ensuremath{N_p}}}   
\newcommand{\Ric}{{\ensuremath{\bf R}}}     
\newcommand{\Nric}{{\ensuremath{N_r}}}   
\newcommand{\ric}{{\ensuremath{r}}}         

\newcommand{\naf}{\mbox{~\textbackslash+~}}

\begin{document}
\bibliographystyle{acmtrans}

\long\def\comment#1{}

\title[Logic\--Based Decision Support for Strategic Environmental Assessment]{Logic\--Based Decision Support for Strategic Environmental Assessment}
\author[M. Gavanelli, F. Riguzzi, M. Milano and P. Cagnoli]
{MARCO GAVANELLI, FABRIZIO RIGUZZI \\
ENDIF - Universit\`a di Ferrara\\
\email{(marco.gavanelli,fabrizio.riguzzi)@unife.it}
\and MICHELA MILANO\\
DEIS - Universit\`a di Bologna\\
\email{michela.milano@unibo.it}\\
\and PAOLO CAGNOLI\\
ARPA Emilia-Romagna\\
\email{PCagnoli@arpa.emr.it}
}
\pagerange{\pageref{firstpage}--\pageref{lastpage}}
\volume{\textbf{10} (3):}
\jdate{March 2002}
\setcounter{page}{1}
\pubyear{2002}
\submitted{7 February 2010}
\revised{}
\accepted{20 March 2010}
\maketitle
\label{firstpage}
\begin{abstract}
Strategic Environmental Assessment is a procedure aimed at
introducing systematic assessment of the environmental effects of
plans and programs. This procedure is based on the so\--called
coaxial matrices that define dependencies between plan activities
(infrastructures, plants, resource extractions, buildings, etc.)
and positive and negative environmental impacts, and dependencies
between these impacts and environmental receptors. Up to now, this
procedure is manually implemented by environmental experts for
checking the environmental effects of a given plan or program, but
it is never applied during the plan/program construction. A
decision support system, based on a clear logic semantics, would
be an invaluable tool not only in assessing a single, already
defined plan, but also during the planning process in order to
produce an optimized, environmentally assessed plan and to study
possible alternative scenarios. We propose two logic\--based
approaches to the problem, one based on Constraint Logic
Programming and one on Probabilistic Logic Programming that could
be, in the future, conveniently merged to exploit the advantages
of both. We test the proposed approaches on a real energy plan and
we discuss their limitations and advantages.
\end{abstract}

\section{Introduction}
\label{sec:intro} Computational Sustainability
\cite{CompSustCP2009} is a very recent, interdisciplinary research
field that aims to apply techniques from computer science,
information science, operations research, applied mathematics, and
statistics to the problem of balancing environmental, economic, and societal
needs for sustainable development.

Among the many possible applications of information
technology to sustainable development, decision
support systems represent a very important topic. Currently,
environmental experts take decisions, perform evaluations and build
plans manually, simply relying on experience, with little or no
support from automated tools.

We believe the main reason why decision support systems are not
widely applied in this field is twofold: first, despite significant
advances in algorithmic research, the current state of decision
support systems still faces severe difficulties or cannot cope at
all with the highly complex structure of sustainability problems.
Second, there is a lack of appropriate models for sustainability
related applications. These models should be developed in tight
collaboration between computer scientists and environmental
scientists, economists and biologists that can provide not only
models and data, but also feedback on system solutions.

Computational Logic can play a very important role in the design
and implementation of decision support systems in this setting.
First it enables a very intuitive and expressive representation of
reality, and second it provides a number of reasoning mechanisms
that can be successfully applied to the many aspects of
sustainability problems: logical inference, constraint reasoning and
probabilistic reasoning.
In addition, Computational Logic tools
rely on a well\--defined semantics, and one can reason on the program to
give explanations of the obtained results (or failure).

Sustainable development encompasses three pillars: society,
economy and the environment. In this paper, we focus particularly
on the environment. We address the problem of defining a logic\--based decision support system for Strategic Environmental
Assessment (SEA), a legally enforced procedure aimed to introduce
systematic evaluation of the environmental effects of plans and
programs. It typically applies to development, waste, transport,
energy and land use plans, both regional and local, within the
European Union. In this paper, we consider as a case study the
assessment of Emilia Romagna regional plans.
 SEA is based on the so\--called {\em coaxial
matrices} that quantify dependencies between activities (e.g.
infrastructures and plants) contained in a plan and positive and
negative environmental impacts (e.g., alteration of woods, water
pollution), and dependencies between impacts and environmental
receptors (e.g., quantity of $CO_2$ in the atmosphere).

We propose two alternative logic\--based approaches: one exploits
Constraint Logic Programming on Real Numbers \clpr, and models
coaxial matrices as sets of linear equations and inequations; this
is a simple, efficient model, that presumes the available
information to be precise, and assumes that influences can be
summed up. The second approach is based on Logic Programs with
Annotated Disjunction (LPADs) where activities and impacts are
combined using the laws of probability. We apply the two
approaches on coaxial matrices referring to eleven types of plans
that legally require the SEA. Experiments are performed on a real
energy regional plan.

The structure of the paper is as follows: section \ref{SEA}
describes regional planning and Strategic Environmental
Assessment along with coaxial matrices. Section \ref{background}
recalls the main concepts behind constraint logic programming and
probabilistic logic programming along with its causal
interpretation. Section \ref{sec:clpr} shows the implementation of
the coaxial matrices in \clpr\, while section \ref{subsec:PLPappli}
describes the approach based on LPAD. Section \ref{exper} presents
experimental results on a real energy plan. A discussion and a
description of open issues conclude the paper.

\section{Strategic Environmental Assessment}
\label{SEA} Regional planning is the science of efficient placement
of land use activities and infrastructures for the sustainable
growth of a region.
Our case study is the Emilia Romagna region of Italy, and we
developed this work on real data provided by the environmental
regional agency. Regional plans are classified into {\em types};
the SEA is legally required for eleven types of plans (namely
Agriculture, Forest, Fishing, Energy, Industry, Transport, Waste,
Water, Telecommunication, Tourism, Urban and Environmental plans),
those addressed in this work. Each plan defines activities that
should be carried out during the plan's implementation. Activities
are roughly divided into six types:

\begin{itemize}
\item infrastructures and plants;

\item buildings and land use transformations;

\item resource extraction;

\item modifications of hydraulic regime;

\item industrial transformations;

\item environmental management.
\end{itemize}
Before any implementation, these plans have to be environmentally
assessed, under the {\em Strategic Environmental Assessment
Directive}. SEA is a method for incorporating environmental
considerations into policies, plans and programs that is
prescribed by European Union policy.


One of the instruments used for assessing a regional plan in Emilia
Romagna are  the so\--called {\em coaxial matrices}, that are a development of the network method \cite{Sorensen}.

One matrix \Mop\ defines the dependencies between the above\--mentioned activities contained in a plan and positive and negative
{\em impacts} (also called {\em pressures}) on the environment. Each element $\mop^i_j$ of the
matrix \Mop\  defines a qualitative dependency between the
activity $i$ and the negative or positive impact $j$. The dependency
can be {\em high}, {\em medium}, {\em low} or {\em null}.
Examples of negative impacts are
energy, water and land consumption, variation of water flows, water
and air pollution and so on. Examples of positive impacts are
reduction of water/air pollution, reduction of greenhouse gas
emission, reduction of noise, natural resource saving, creation of
new ecosystems and so on.

The second matrix \Mpr\ defines how 
the impacts influence
environmental receptors.
Each element $\mpr^i_j$ of the matrix \Mpr\ defines a qualitative dependency between the negative or
positive impact $i$ and an environmental receptor $j$. Again the
dependency can be {\em high}, {\em medium}, {\em low} or {\em null}. Examples of
environmental receptors are the quality of surface water and
groundwater, quality of landscapes, energy availability, wildlife
wellness and so on.

The matrices currently used in Emilia Romagna  contain 93
activities, 29 negative impacts, 19 positive impacts and 23
receptors and refer to the above\--mentioned 11 plans.

The coaxial matrices are currently used by environmental experts
that manually evaluate a single, already defined, plan. A plan
basically defines the so\--called {\em magnitude} of each activity:
magnitudes are real values that intuitively express {\em ``how much''}
of an activity is performed. The unit is different for each activity: for example,
for activity {\em Thermoelectric power plants} the magnitude says how many MW
of electric power will be produced by thermoelectric plants,
while the magnitude of {\em Oil/gas/steam pipelines} gives the number of kilometers
of pipes installed.
A manual evaluation of alternatives and {\em what-if} queries are
very difficult to consider. In addition, planning is now carried
out without a rigorous consideration of environmental aspects
contained in the coaxial matrices.

 In this paper, we propose two logic\--based approaches for
the design of a decision support system that can be used  to assess
a single, already defined plan, to evaluate different scenarios
during the planning phase or to optimize the definition of land use
activities and infrastructures.

In both cases, we convert the qualitative values into real numbers in the interval $[0,1]$.
The environmental expert suggested the values to be 0.25 for {\em low}, 0.5 for {\em medium},
and 0.75 for {\em high}.

The first approach is based on Constraint Logic Programming on
Real numbers (\clpr), that is extremely efficient when dealing
with linear equations. On the other hand, this approach does not
take into consideration the subjective and stochastic nature of
the available data: each value in the matrices
is simply used as a coefficient in a linear equation, so we assume
that positive and negative impacts derived from planned activities
can be summed. While in general, impacts can indeed be summed, in
some cases a mere summation is not the most realistic relation and
more sophisticated combinations should be considered.

For this reason, we also evaluate a Causal Probabilistic Logic
Programming approach that is grounded on the well\--established
theory of probability and causality. The same coefficients are now
interpreted as probabilities, that will be combined through 
probability laws to provide the likelihood of a given receptor
being affected. The price to be paid is a higher computation time.
A realistic decision support system should merge the two
approaches and this is a subject of the current research activity.

\section{Background}
\label{background} We provide some preliminaries on the two
logic\--based techniques used in this paper.

\subsection{Constraint Logic Programming}
\label{sec:intro_clpr}

Constraint Logic Programming (CLP) \cite{JM94} is a class of
programming languages which extend classical Logic Programming.
Variables can be assigned either terms (as in Prolog), or
interpreted values, taken from a {\em sort}, that is a parameter
of the specific CLP language. For example, we can have \clpr\
\cite{clpr}, on the sort of real values, or CLP(FD), in which
variables range on finite domains. The sort also contains
interpreted functions (that, in numerical domains, can be the
usual operations +, -, $\times$, etc.) and predicates (e.g., $<$,
$\neq$, $\geq$, etc.), which are called {\em constraints}. The
declarative semantics gives the intuitive interpretation of the
specific sort to constraints and interpreted terms: e.g.,
$1.3+2<5$ is {\em true} in \clpr. The operational semantics
resembles that of Prolog for atoms built on the usual predicates
(i.e., those predicates defined by a set of clauses), but stores
the interpreted ones, the constraints, to a special data
structure, called the {\em constraint store}. The store is then
interpreted and modified by an external machinery, called the {\em
constraint solver}. The solver is able to check if the conjunction
of constraints in the store is (un)satisfiable, and is also able
to modify the store, possibly simplifying it to a refined state.
Usually, the constraint solver does not perform {\em complete}
propagation: if it returns {\em false}, then there is definitely
no solution, but in some cases it may fail to detect infeasibility
even if no solution exists.

\clpr\ is the instance of CLP in which variables range on the reals.
The available constraints are linear equalities and
inequalities, and the solver is usually implemented through the
simplex algorithm, which is very fast and is complete for linear
(in)equalities (it always returns {\em true} or {\em false}). 
Also, the user can communicate an {\em objective function} to the
solver: a linear term that should be
minimized or maximized while satisfying all constraints.

Many implementations  of \clpr\ exist nowadays \cite{KoninckSD06}, and many Prolog
flavours \cite{B-Prolog,Ciao_Prolog} have their own \clpr\ library. We decided to adopt
\eclipse, that features a library called {\em Eplex} \cite{eplex}.
This library interfaces \eclipse\ to an external mixed integer
linear programming solver, which can be either a state-of-the-art
commercial one (like CPLEX or Xpress-MP), or an open source
solver. By default, Eplex hides most of the details of the solver,
but nevertheless, when required, the user can trim various
parameters to boost the performance, and also inspect the
internals of the solver. This feature becomes very useful in
practical applications, and will be used to provide additional
valuable information to the user, as detailed in
Section~\ref{sec:clpr}.

\subsection{Causal Probabilistic Logic Programming}
\label{sec:prob} In this section we first present Probabilistic
Logic Programming and then we discuss how to model causation with
it.
\subsubsection{Probabilistic Logic Programming}
\label{subsec:plp}

The integration of logic and probability has been widely studied in Logic Programming and various languages semantics have been proposed, such as Probabilistic Logic Programs \cite{DBLP:conf/lpar/Dantsin91},  Independent Choice Logic \cite{DBLP:journals/ai/Poole97}, PRISM \cite{DBLP:conf/ijcai/SatoK97}, pD \cite{DBLP:journals/jasis/Fuhr00}, CLP(BN) \cite{SanPagQaz03-UAI-IC} and ProbLog \cite{DBLP:conf/ijcai/RaedtKT07}.

Logic Programs with Annotated Disjunctions (LPADs) \cite{VenVer04-ICLP04-IC}
are particularly suitable for reasoning about causes and effects \cite{DBLP:journals/tplp/VennekensDB09}. They extend logic programs by allowing  clauses to be disjunctive and by annotating each atom in the head with a probability. A clause can be causally interpreted by supposing that  the truth of the body causes the truth of one of the atoms in the head non-deterministically chosen on the basis of the annotations.

An LPAD theory $T$ 
consists of a finite set of
 \emph{annotated disjunctive clauses}. These clauses have the following form
$$(H_1:\alpha_1)\vee(H_2:\alpha_2)\vee \ldots\vee (H_h:\alpha_h):- B_1,
B_2, \ldots B_b$$
where the $H_i$s are logical atoms, the $B_i$s are logical
literals and the $\alpha_i$s are real numbers in the interval $[0,1]$
such that $\sum_{i=1}^h \alpha_i\leq 1$. If $\sum_{i=1}^h \alpha_i<1$, the head of the clause implicitly contains an extra atom $null$ that does not appear in the body of any clause and whose annotation is $1-\sum_{i=1}^h \alpha_i$. If $C$ is the clause above, $H(C,i)$ is $H_i$, $\alpha(C,i)$ is $\alpha_i$ and $body(C)$ is $B_1,
B_2, \ldots B_b$.

The semantics of a non-ground theory $T$ is defined through its grounding $g(T)$ and 
\citeN{VenVer04-ICLP04-IC} require that $g(T)$ is finite.

An \emph{atomic choice} $\chi$ is a triple  $(C,\theta,i)$ where $C\in T$, $\theta$ is a substitution that grounds $C$ and $i\in\{1,\ldots,n\}$ where $n$ is the number of atoms in the head of $C$. $(C,\theta,i)$  means that, for the ground clause $C\theta$, the head $H(C,i):\alpha(C,i)$ was chosen. A \emph{selection} $\sigma$ is a set of atomic choices such that for each clause $C\theta$ in  $g(T)$ there exists one and only one atomic choice
$(C,\theta,i)$ in $\sigma$. We denote the set of all
selections   of a program $T$ by $\mathcal{S}_T$.

A selection $\sigma$ identifies a normal logic program $T_\sigma=\{(H(C,i):- body(C))\theta|$ $(C,\theta,i)\in \sigma\}$ that is called an \emph{instance} of $T$. A probability distribution is defined over the space  of instances by assuming
independence among the choices made for each clause, thus the probability $P_\sigma$ of an instance $T_\sigma$ is given by
$P_\sigma=\prod_{(C,\theta,i)\in \sigma}\alpha(C,i)$.

The meaning of the instances of an LPAD is given by the well\--founded semantics.
For each instance $T_\sigma$, we require that its well\--founded model $WF(T_\sigma)$ is total, since we want to model uncertainty only by means of disjunctions.
%
%
%
%

The probability of  a  formula $Q$ is given by the sum of the probabilities of the instances in which the formula is true according to the well\--founded semantics:
$$P(Q)=\sum_{\sigma\in  \mathcal{S}_T,WF(T_\sigma )\models Q} P_\sigma$$
%
%
%
%
%
%
%
An LPAD $T$ can be translated into a Bayesian network $\beta(T)$
that has a Boolean random variable for each ground atom plus a
random variable $choice_{C\theta}$ for each grounding $C\theta$ of
each clause $C$ of $T$ whose values are the atoms in the head of $C\theta$ plus $null$.

$choice_{C\theta}$ assumes value $H(C,i)\theta$  with probability $\alpha(C,i)$ if the configuration of its parents makes the body  true, while it assumes value $null$ with probability 1 if the configuration makes the body false.
The parents of ground atom $A$ are all the $choice_{C\theta}$ variables such that $A$ appears in the head of $C\theta$. $A$ assumes value true with probability 1 if one of the parent choice variables assumes value $A$, otherwise it assumes value false with probability 1.

Various approaches have been proposed for computing the
probability of queries from an LPAD. Riguzzi
\shortcite{Rig08-ICLP08-IC} discusses an extension of SLG
resolution, called SLGAD, that is able to compute the probability
of queries by repeatedly branching on disjunctive clauses. A
different approach was taken by Meert et al.
\shortcite{MeeStrBlo08-ILP09-IC}, where an LPAD is first
transformed into its equivalent Bayesian network and then
inference is performed on the network using the variable
elimination algorithm. Riguzzi \shortcite{Rig-AIIA07-IC} presents
the \texttt{cplint} system that first finds explanations (sets of
atomic choices) for queries and then computes the probability by
means of Binary Decision Diagrams, as proposed in
\cite{DBLP:conf/ijcai/RaedtKT07} for the ProbLog language.
 \texttt{cplint} was used in the experiments in Section~\ref{causal-ex} because of its speed \cite{Rig09-LJIGPL-IJ}.

\subsubsection{Causal Models} \label{causal-models} Determining
when an event causes another event is very important in many
domains, take for example science, medicine, pharmacology or
economics. Causality has been widely debated by philosophers and
statisticians: often it has been confused with correlation, while
they are in fact distinct concepts, since two events may be
correlated without one causing the other.  Recently,
Pearl  \shortcite{Pea00-book} helped to clarify the concept of causation by
discussing how to represent causal information and how to perform
inference from it. He illustrates two types of
causal models: causal Bayesian networks and structural equations.

Causal Bayesian networks differ from standard Bayesian networks because the edge from variable $X$ to variable $Y$ means that $X$ is a cause for $Y$, while in standard Bayesian networks it simply means that there is a statistical dependence.

Pearl \shortcite{Pea00-book}  proposed an approach for computing the probability of effects of actions and suggested to use the notation  $P(y|do(x))$ to indicate the effect of the 
action of setting the variable $X$ to value $x$ on the event of variable $Y$ taking value $y$.
$P(y|do(x))$ is different from the probability of $y$ given $x$ ($P(y|x)$) because we do not simply observe $X=x$ but we intervene on the model by making sure $X=x$ is true.

The technique proposed in \cite{Pea00-book} for computing $P(y|do(x))$ consists of removing the parents of $X$ from a causal Bayesian network, setting $X$ to $x$ and computing  $P(y)$ in the obtained network.

This approach can be applied to probabilistic logic languages that can be translated to Bayesian networks, such as LPADs. In order to compute the probability  of a ground atom $Y$ of being true given an intervention that consists of making a ground atom $X$ true from an LPAD $T$,  we need to  remove $X$ from the head of all the clauses that contain it and add $X$ as a fact to $T$.
The probability of $Y$ can then be computed from
the resulting LPAD $T'$ by using standard inference, i.e., by computing  $P(Y)$.

\section{Coaxial Matrices in CLP(R)}
\label{sec:clpr}

The coaxial matrices can be simply interpreted as a linear programming model.
Amongst the many ways to invoke a linear programming solver,
we decided to use \clpr; in this way the model is written as a knowledge base in a computational logic language,
that could be 
easier to integrate with the probabilistic approach in Section~\ref{subsec:PLPappli}.



In more detail,  the environmental impacts caused by 
activity $i$ (with magnitude $\ope_i$) can be estimated with the system of linear
equations
$$
 \forall j \in \{1,\dots,\Npre\} \quad \pre_j = \mop^i_j \ope_i.
$$
When considering a whole regional plan, we sum up the contributions of all the activities and obtain the estimate of the
influence on each environmental impact:
\begin{equation}
 \forall j \in \{1,\dots,\Npre\} \quad  \pre_j =\sum_{i=1}^\Nope \mop^i_j \ope_i.
\label{eq:pre_ope}
\end{equation}
In the same way, given the vector of environmental impacts $\Pre=(\pre_1,\dots,\pre_\Npre)$, one can estimate the
influence on the environmental receptor $\ric_i$
by means of the matrix \Mpr, that relates impacts with receptors:
\begin{equation}
 \forall j \in \{1,\dots,\Nric\} \quad \ric_j = \sum_{i=1}^\Npre \mpr^i_j \pre_i.
\label{eq:ric_pre}
\end{equation}
The system of equations (\ref{eq:pre_ope}-\ref{eq:ric_pre}) are imposed as constraints in a \clpr\ program;
thanks to this formalisation, a number of queries of high interest both for the
planner and for the evaluator of the environmental policy
can be posed to the system as \clpr\ goals.

The final goal for the evaluator of the environmental policy is
computing the environmental footprint of a devised plan. The plan is
given as a set of values representing the magnitude of each of the
activities. 
In other words, given the
set of values $\Ope = (\ope_1,\dots,\ope_\Nope)$, we can compute the
environmental footprint $\Ric = (\ric_1,\dots,\ric_\Nric)$, simply
by applying equations (\ref{eq:pre_ope}) and
(\ref{eq:ric_pre}).

Another query studies the impact of a single unit (in a
standardized format) of activity $\ope_i$; for example, we are
interested to know what the environmental footprint is of
producing 1 MW of electric power through a thermoelectric plant. We
instantiate the vector of activities to a unary vector with
$\ope_{i} = 1$ if $i=therm$ and $\ope_{i}=0$ otherwise:
$$\Ope = (0,0,\dots,1,\dots,0)$$
In this way, one can find out, by looking at the resulting vector
\Ric, which of the receptors are (positively or negatively)
influenced by the devised activity. Also, one can  get an
estimate of those receptors that are more heavily influenced, and
those that are only marginally influenced. This query can also be used
by experts to calibrate the numbers in the coaxial matrices, by
considering each activity singularly.

Another important query for the final user 
is asking
which of the possible activities (always in normalized form) has a
major impact on some given receptor $\ric_i$. In fact, in \clpr,
one can maximize or minimize some objective function, so the model
becomes
$$\begin{array}{cc}
    &   max(\ric_i) \\
s.t. & (\ref{eq:pre_ope}) (\ref{eq:ric_pre}) \\
    & \sum_j \ope_j =1 \\
    & \forall j, \ope_j \ \mbox{is integer}
\end{array}
$$
Finally, if there are laws imposing limits on some receptors
(limits for CO$_2$ emissions, for example) one can very easily
impose constraints on receptors (e.g., $\ric_{CO_2} \leq
limit_{CO_2}$), and find if an activity can either be performed at
all, or if it requires some compensation (e.g., another activity
that improves on the receptor, like reforestation for $CO_2$), or
if it can be done in association with other activities.

In cases where there are two or more alternative activities that
cater for the same need, the regulations prescribe that
alternatives should be studied, and compared. For example, the
need for additional electrical power is satisfied by building a
new plant; however one can choose the type of plant, depending on
the environmental conditions. In an area with highly polluted air,
a thermoelectric plant could raise the pollution over the law
limit, so a different type of plant could be devised, like a solar
power plant. On the other hand, a solar plant could be too
expensive, and make other activities that are
necessary in the area (e.g., building a school, a hospital, etc.) unaffordable.
In this case, the planner can impose a constraint stating that
there is a regional need for at least $k$ MW of electrical power;
he/she imposes
$$\sum_{i\in PowerPlants} \ope_i \geq k$$
(where $PowerPlants$ is the set of indices in the vector \Ope\
corresponding to those plants that provide electrical power) and
then can optimize for one of the receptors, e.g., $r_{CO_2}$, or
some weighted sum of receptors of interest. Or, the planner may
ask what the maximum power is that can be generated in the region
without violating the law limits on the receptors  
$$\begin{array}{cc}
    &   max\sum_{i\in PowerPlants} \ope_i \\
s.t. & (\ref{eq:pre_ope}) (\ref{eq:ric_pre}) \\
\forall i \in \{1,\dots,\Nric\} & \ric_i \leq limit_i \\
\end{array}
$$
In this way, we find the maximum number of MW that can be
produced, as well as the electrical power produced by each type of
plant. Note that in this way the solver could find an assignment
that imposes the execution of compensation activities, as hinted
earlier. If there are not enough resources for compensation, we
can impose that such activities must not be performed (e.g., by
assigning value 0 to all these activities), or we can impose that,
given a vector {\bf C} with the cost of each activity, the total
cost of the activities should not be higher than the allotted
finances $F$:
\begin{equation}
    \sum_{i=1}^\Nope c_i \ope_i \leq F
\label{eq:cost_limit}
\end{equation}
In the same way, other types of resources, like time, person-months, energy, can be taken into consideration.

We are currently improving the model to take into account the fact
that different activities can have different impacts on the
environment depending on the type of zone they are placed. For
example, if we build a power line within a natural park, its
impact is definitely higher than building it near a city. An
additional feature we are studying is the fact that, depending on
the zone we are considering, different receptors might have
different weights. For instance, the water quality is extremely
important on a river delta, where the whole ecosystem relies on the river water,
while it is less important in an
industrialized area.

\subsection{Sensitivity Analysis}
\label{sec:sensitivity}
The simplex algorithm provides the optimal value of the objective
function, the optimal assignment to the decision variables, and also
other information that is of high interest for the decision maker.
In particular, it provides the so-called {\em reduced costs}, and
the {\em dual solution}. These indicators provide precious
information on the sensitivity of the found solution to the
parameters of the constraint model.

The dual solution is a set of values that correspond to the
constraints. It can be thought of as the derivatives of the objective
function with respect to the right hand side (RHS) of the
constraints. This means that we immediately see, in the dual
solution, which of the constraints are {\em tight}, i.e., which
would change the value of the objective function if the RHS
coefficient changes. For example, if we are optimizing the number
$f_{MW}$ of MW of electric power and we have a constraint
$\ric_{CO_2} \leq limit_{CO_2}$, the corresponding dual value
$d_{CO_2}\frac{\partial f_{MW}}{\partial limit_{CO_2}}$ in the optimal solution answers the
question: {\em ``How much would the production of energy decrease
in case the limit of $CO_2$ lowers one point?''}
This is 
important information, since regulations change, and tend to become more strict.

The same analysis can be performed on the problem of optimizing
some (weighted sum of) receptors, given a total number of plants
(or required MW). In this case, the dual value associated to a
constraint represents how much the receptor will improve if that
constraint is partially relaxed (if the RHS becomes less strict).
For example, suppose we are optimizing the emissions of nitrogen
oxides ($NO_x$), and we have the constraint (\ref{eq:cost_limit})
stating a limit on the total cost of the activities, for example,
in euro. After obtaining the optimal value, the planner could ask:
{\em ``Suppose now that we had more money: if I add one euro, how
much would the emissions of $NO_x$ decrease?''} The answer is the
dual value $d_e$ of the constraint (\ref{eq:cost_limit}).
This analysis is
very attractive for the evaluator.

\section{A  Causal Model for Coaxial Matrices}
\label{subsec:PLPappli}

In this section we consider an interpretation of Coaxial Matrices that differs from the one in Section \ref{sec:clpr}. Instead of associating a real number to each activity, impact and receptor, we associate a Boolean random variable to each of them and we consider the interaction levels expressed in the matrix as probabilistic causal dependencies. In this approach, we assume that an activity is either carried out or not, an impact is either present or not  and a receptor is either achieved or not. In other words, we do not consider the magnitude or level of the variables under analysis. We used this approximation to get useful insights on the probabilistic modeling of the problem. In the future, we plan to consider more refined approximations with multivalued random variables or even continuous random variables.

Activities, impacts and receptors are  represented by LPAD atoms (propositions) and the effects of activities on impacts and of impacts on receptors are expressed by means of LPAD rules that represent the Coaxial Matrices.

The model thus contains rules that express the effect of the activities on the negative impacts (where  $\mop_j^i$ is an element of the $\Mop$ matrix):
$$negative\_impact_j:\mop_j^i \mbox{~:-~} activity_i.$$
%
%
Also, there are rules expressing the effect of the activities on the positive impacts:
$$\mathit{positive\_impact}_j:\mop_j^i \mbox{~:-~} activity_i.$$


For example, the model contains the rule
\begin{verbatim}
'Dispersion of dangerous materials':0.75 :-
  'External movements of dangerous materials'.
\end{verbatim}
that relates  an activity and a negative impact, and the rule
\begin{verbatim}
'Creation of work opportunities':0.5 :-
  'External movements of dangerous materials'.
\end{verbatim}
that relates an activity and a positive impact.

Negative impacts reduce the probability of receptors, while positive impacts increase it. 
However adding a clause with a certain atom in the head can only increase the probability of the atom.
To model the fact that negative impacts lower the probability of receptors,  we use, for each receptor $receptor_k$, two auxiliary predicates $receptor\_pos_k$ and $receptor\_neg_k$ that  collect the evidence in favor or against the achievement of the receptor. 

The  rules that express the negative effect of the negative impacts on the receptors take the form:
$$receptor\_neg_k:\mpr_k^j \mbox{~:-~} negative\_impact_j.$$
%
while the rules that express the positive effect of the positive impacts on the receptors take the form:
$$receptor\_pos_k :\mpr_k^j \mbox{~:-~} \mathit{positive\_impact}_j.$$
%
where  $\mpr_k^j$ is an element of the $\Mpr$ matrix.
For example, the rule
\begin{verbatim}
'Human health/wellbeing_neg':0.25:-
  'Dispersion of dangerous materials'.
\end{verbatim}
expresses a negative effect of a negative impact on a receptor, and the rule
\begin{verbatim}
'Human health/wellbeing_pos':0.75:-'Creation of work opportunities'.
\end{verbatim}
expresses a positive effect of a positive impact on a receptor.

Finally, the positive and negative evidence regarding the receptor are combined with the following rules:
$$\begin{array}{llll}
receptor:0.1 & \mbox{:-} & \naf receptor\_pos,  receptor\_neg.\\
receptor:0.5 & \mbox{:-} & \naf  receptor\_pos,  \naf  receptor\_neg.\\
receptor:0.5 & \mbox{:-} & receptor\_pos,  receptor\_neg.\\
receptor:0.9 & \mbox{:-} & receptor\_pos,  \naf  receptor\_neg.
\end{array}$$

\noindent
For example, the model contains the rules
\begin{verbatim}
'Human health/wellbeing':0.1 :- \+ 'Human health/wellbeing_pos',
                                   'Human health/wellbeing_neg'.
'Human health/wellbeing':0.5 :- \+ 'Human health/wellbeing_pos',
                                \+ 'Human health/wellbeing_neg'.
'Human health/wellbeing':0.5 :-    'Human health/wellbeing_pos',
                                   'Human health/wellbeing_neg'.
'Human health/wellbeing':0.9 :-    'Human health/wellbeing_pos',
                                \+ 'Human health/wellbeing_neg'.
\end{verbatim}
that collect positive and negative effects on the receptor {\em ``Human health/wellbeing''}.

These rules express the fact that {\em ``Human health and wellbeing''} is unlikely if there is no positive evidence on it and there is negative evidence on it (first rule). It is very likely if there is positive evidence on it and no negative evidence on it (last rule). In the other cases, the probability of {\em ``Human health and wellbeing''} is in between (second and third rule).

All the parameters were subjectively estimated and validated by the expert.

\section{Experiments}
\label{exper}

\subsection{Experimental results of CLP(R)}
\label{sec:experiments_clpr}

The agency for the environment of the Emilia-Romagna region
(Italy) kindly provided us with the coaxial matrices used for
assessing eleven types of plans (that we translate into the CLP
model) and the data of a regional energy plan: for each of the
activities, we have a {\em ``magnitude''} value. Thanks to the CLP
model described earlier, we are able to compute the
corresponding values of impacts and receptors.

Initially the results were counterintuitive: the
considered plan concerned energy (aimed at raising the available
electrical power in the region), while the receptor {\em energy
availability} had a lower value than the previous year. These
types of results may be partially due to the qualitative
information contained in the matrices, but also highlight
possible human mistakes in the data of the matrices. Indeed, a
flaw was found (and fixed) in the matrices, showing how
logic-based decision support can contribute
to increase the reliability of 
the environmental assessment.

Once the human mistakes had been corrected, we reran the experiments.
The new results were highly
appreciated by the evaluator: the decision support system foresaw
strong decrease of {\em quality of air} (mainly due to the boost
on thermoelectric plants), and {\em water availability} (since
thermoelectric plants need refrigeration).

As the plan had a large impact on some receptors, we tried to
improve it from an environmental viewpoint: the magnitude of each
of the activities was allowed to deviate up to 1\% with respect to
the original plan, and we optimized the {\em quality of air}
receptor. We had an improvement of about 20.3\% on this receptor,
which shows that even by allowing small variations one can get
significant improvements. On the other hand, we had a decrease of
industrial indicators, such as the {\em availability of productive
resources} or the {\em availability of energy}.

We also tried two dual goals.
The first considers the given plan, keeps all activities constant except the building of (various types of) power plants,
fixes the amount of produced energy, and tries to optimize on the quality of air.
The second, instead, maximizes the  electrical power supply without sacrificing any of the
environmental receptors, i.e., none of the receptors could worsen with respect to the original plan.
The first query gave a positive result: by producing electricity with environmental friendly power
plants (wind-powered aerogenerators) we could produce the same amount of energy
but have a 57\% improvement on the quality of air.

The second, instead, had a negative result: we could not improve the produced electrical power
without worsening at least one receptor.
These seemingly contradictory results actually have an interesting explanation.
The receptors taken into account by the environmental assessment
range on all aspects influenced by a human activity,
spanning, e.g., from {\em value of cultural heritage} to {\em stability of riverbeds},
from {\em quality of underground water} to {\em visual impact on the landscape}.
Aerogenerators, recommended in the previous optimization, have a significant visual impact,
so they are not implementable unless we relax the visual requirement.

Computing time for  this analysis was hardly measurable:
all times were far less than a fraction of second
on a modern PC.
Thanks to such a fast computation, we could  comment the results of the queries online with the experts of the regional agency,
identify errors in the provided data,
and try variations of the parameters.

\subsection{Causal Model}
\label{causal-ex}
Given the causal model presented in Section~\ref{subsec:PLPappli}, we can ask various what\--if queries 
\begin{enumerate}
\item if these activities are performed, what is the probability of a certain impact of appearing? \label{inf-queries}
\item if these works are performed, what is the probability of a certain receptor 
being satisfied?\label{rec-queries}
\end{enumerate}
Queries of type \ref{rec-queries} are more interesting because they relate the works directly with their final effects of interest. However, they are also more complex to compute.
Moreover, the queries above can be generalized to the case in which the activities are performed with a certain probability.

We can answer the queries above by following the approach described in \ref{causal-models}: we add a fact for each activity that is carried out and we ask for the probability of the query from the modified program.

We report on a number of queries 
together with their execution times on Linux machines with an Intel Core 2 Duo E6550 (2333 MHz) processor and 4 GB of RAM.


The probability of the negative impact {\em ``Dispersion of Dangerous Materials''} performing the activities {\em ``External movements of dangerous materials''} and {\em ``Internal movements of dangerous materials''} is 0.937500. The CPU time was below $10^{-6}$ seconds. 

The probability of the receptor {\em ``Human health/wellbeing''} given that we perform the activities {\em ``External movements of dangerous materials''} and {\em ``Internal movements of dangerous materials''} is 0.546915 and the query took 22.713
 seconds. 

If we perform the activity  {\em ``Industrial processing and transformation''} the probability of the receptor {\em ``Human health/wellbeing''}  is 0.474918, computed in 84.453
 seconds. 
This query takes longer than the previous ones because the work {\em ``Industrial processing and transformation''} has an influence on many more impacts than {\em ``External movements of dangerous materials''} and {\em ``Internal movements of dangerous materials''} and all these influences must be combined to find the effect on {\em ``Human health and wellbeing''}. To give the reader an idea of the complexity of this query, there are 655,660 explanations, 12,847,036 atomic choices appear in the explanations and 42 random variables are involved.

The probability of  receptor {\em ``Atmosphere quality, microclimate''} given the action {\em ``Industrial processing and transformation''} is 0.360851. 
The CPU time was 0.02s. 

If we add the activity {\em ``Oil and gas extraction plants''} the probability of the receptor {\em ``Atmosphere quality, microclimate''} lowers to 0.326481, computed in 6.852s

By adding the activity {\em ``Fire extinguishing plants''} the probability  of the receptor {\em ``Atmosphere quality, microclimate''} rises to 0.454471, due to the positive effects of the last activity. The CPU time was 92.67
 seconds. 

As can be seen from the last three cases, increasing the number of activities increases the  computation time, since we have to combine the effects of the different causes. The last query has
606,726 explanations,
10,973,022 atomic choices appear in the explanations and 36 random variables are involved.

\subsection{Comparison}
\label{sec:comparison}

\begin{figure}
\includegraphics[width=9cm]{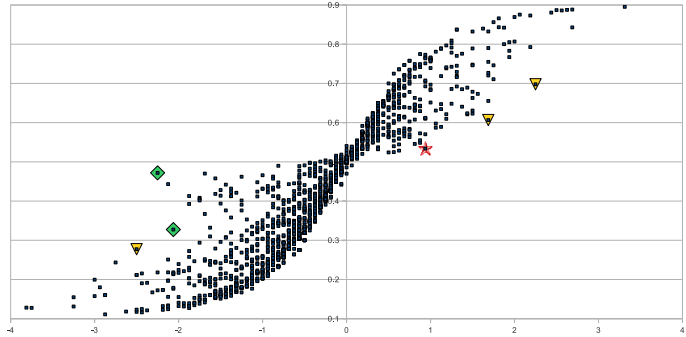}
\caption{Scatter plot of linear vs probabilistic values}
\label{fig:scatter_plot}
\end{figure}

As hinted earlier, we developed two models with the final aim to integrate the two.
Before such an ambitious goal can be reached, we need to identify strengths and weaknesses of the
two.
In order to have a systematic comparison, we produced two tables (one for each approach)
in which each cell contains the effect of a single activity on a single receptor. One table reports the results of the linear model and the other those of the causal model.
The tables are thus of size $\Nope \times \Nric=93 \times 23=2139$.

In the scatter plot of Figure~\ref{fig:scatter_plot} we draw the results of the causal model against those of the linear model:
the linear model results are on the $x$-axis, while in the causal model results are on the $y$-axis.
As we can see, most of the point are clustered along a simple curve, which seems to indicate a close relationship between the two models that we are going to investigate in the near future. Moreover, whenever the linear value is positive,
the probability of improving the receptor is greater than 0.5 and vice-versa,
showing that the two models may disagree on the values, but they agree on the direction of the effect on the receptor.

To better investigate the results, we considered the points farthest from the curve that are highlighted in Figure~\ref{fig:scatter_plot}.
Since these are the points for which the linear and causal model differ the most,
we asked the expert to evaluate the results for those points, to understand which model gave the best result.
For the points shown as triangles in Figure~\ref{fig:scatter_plot}, the expert was unable
to state which answer is better.
For the points with a diamond symbol, the \clpr\ approach gave a better result.
In the point shown as a star, both approaches failed to give a correct result.


From these results, we can say that often the effects can be summed up,
although in some cases other combinations could be necessary.
In future work, we plan to try other techniques, such as fuzzy logic, and to use 
probabilities to model the uncertainty of the parameters of the matrices.

%
%

\section{Conclusions}
\label{sec:conclusions}

%
%

The environmental assessment is now becoming a systematic
procedure imposed by the laws, and its importance is doomed to
increase year after year. In this paper, we proposed how two
technologies taken from computational logic can successfully
address practical problems of the environmental assessment. The
work was conducted using the real data used in previous years for
the environmental assessment of Emilia Romagna plans.

Constraint Logic Programming showed an efficient management of
large models and provides useful sensitivity analysis that can be
used by planners and evaluators to assess multiple alternative
scenarios. The drawback is that contributions from different
activities and different impacts are merely summed, while in some
cases more sophisticated combinations are required.
We plan to investigate the use of other CLP languages, like CLP(FD),
that allow for more general constraints.

On the other hand, the probabilistic model takes into
consideration the subjective and stochastic nature of the provided
data, paying the cost of a higher computational effort. In
addition, some activity contributions and impacts should indeed be
summed, while others are conveniently merged through probability
laws.

We believe that computational logics can have a big impact in this field.
One future work will be trying to merge the two models into a single component:
in CL and AI, formalisms have been proposed that take 
constraints and probabilities under a same umbrella, like the Valued CSP model \cite{ValuedCSP}
or the semiring framework \cite{semiring}.

%

\bibliography{odessa,bib,lpadlocal}

\end{document}